\documentclass{article}

\usepackage{iclr2027_conference,times}
\iclrfinalcopy

\usepackage[utf8]{inputenc}
\usepackage[T1]{fontenc}
\usepackage{courier}
\usepackage{url}
\usepackage{booktabs}
\usepackage{amsfonts}
\usepackage{amsmath}
\usepackage{amssymb}
\usepackage{mathtools}
\usepackage{amsthm}
\usepackage{nicefrac}
\usepackage{microtype}
\usepackage{xcolor}
\usepackage[normalem]{ulem}
\usepackage{pifont}
\usepackage{array}
\usepackage{graphicx}
\usepackage{float}
\usepackage{makecell}
\usepackage{multirow}
\usepackage{bm}
\usepackage{hyperref}

\usepackage{amsmath,amsfonts,bm}

\def\eqref#1{equation~\ref{#1}}

\def\1{\bm{1}}

\DeclareMathAlphabet{\mathsfit}{\encodingdefault}{\sfdefault}{m}{sl}
\SetMathAlphabet{\mathsfit}{bold}{\encodingdefault}{\sfdefault}{bx}{n}

\usepackage{xcolor}

\newcolumntype{C}[1]{>{\centering\arraybackslash}p{#1}}

\title{VascuSR: Iterative Anatomy-Aware Diffusion for CT Super-Resolution}

\author{
\begin{tabular}{c}
\begin{tabular}{ccccc}
Hairong Ni$^{*\dagger 2}$ &
Jiahe Huang$^{*1}$ &
Minghao Guo$^{3}$ &
Wojciech Matusik$^{3}$ &
Rose Yu$^{1}$
\end{tabular}
\\[0.5em]
{\normalfont
$^{1}$University of California, San Diego
\qquad
$^{2}$University of Michigan
\qquad
$^{3}$MIT CSAIL
}
\end{tabular}
}

\begin{document}

\maketitle

\begingroup
\renewcommand{\thefootnote}{}
\makeatletter
\renewcommand{\@makefntext}[1]{\noindent #1}
\makeatother
\footnotetext{
$^{*}$Equal contribution.
\quad
$^{\dagger}$Work partially done at the University of California, San Diego.
\quad
Correspondence to: Jiahe Huang \texttt{<chh118@ucsd.edu>} and
Rose Yu \texttt{<roseyu@ucsd.edu>}.
}
\addtocounter{footnote}{-1}
\endgroup

\begin{abstract}

Generative super-resolution (SR) can recover sharp detail in computed
tomography (CT) images from severely downsampled observations, but the missing high-frequency content is not uniquely determined by the low-resolution input. Consequently, visually plausible reconstructions may contain anatomically unsupported vascular detail, such as missing, spurious, or incorrectly connected vessels.
To overcome this challenge, we introduce \textbf{VascuSR}, an iterative anatomy-aware diffusion framework that combines image reconstruction with explicit 3D vascular guidance.
First, to provide richer anatomical guidance, we use an explicit
3D vascular representation that captures structural information
beyond individual slices. This representation also simplifies vascular refinement by allowing a lightweight
3D generative refiner to focus on vascular structure rather than complex
CT appearance.
Second, to guide slice-wise SR, a lightweight anatomy controller injects the refined structure into the diffusion decoder, while iterative feedback updates this guidance from each new SR volume.
On re-annotated MSD Task08 and HVM datasets, VascuSR outperforms interpolation and learning-based SR baselines in segmentation accuracy (Dice and clDice) and 3D vascular structure (connectivity and mean centerline distance), while maintaining image quality comparable to its SR backbone.
Compared with learning-based baselines, VascuSR reduces class-averaged connected-component counts by over 20\%
on both datasets and mean centerline distance by over 10\% on MSD.


\end{abstract}

\section{Introduction}
\vspace{-0.5em}
Medical image super-resolution (SR) aims to recover image detail from low-resolution observations while
preserving anatomically meaningful structure. This is particularly challenging for contrast-enhanced vascular CT, where vessels are thin, highly branched, and span a wide range of diameters and curvatures: small peripheral branches may additionally exhibit weak contrast against surrounding tissue \citep{lesage2009review}. At high upsampling factors, much of this fine image detail is poorly determined by the observed low-resolution image, making the reconstruction inherently ambiguous~\citep{ledig2017photo}.

Recent deep SR has progressed from deterministic regression to generative
reconstruction, enabling sharper and more realistic image detail
\citep{zhang2018image,liang2021swinir,ledig2017photo,saharia2023image,
yue2023resshift}. However, visual plausibility does not guarantee anatomical fidelity. Errors that are small in image space can be structurally significant: a short discontinuity may disconnect an otherwise continuous vessel tree, while a spurious connection may alter its branching structure. Such failures are not well characterized by conventional image-quality metrics. High SSIM, for example, can coexist with hallucinated anatomical structures in medical reconstruction \citep{muckley2021fastmri}, and improved image-quality metrics do not necessarily translate to better downstream detection performance \citep{li2021assessing}.
Deep-learning-based CT reconstruction may also blur or remove small vessels despite producing
visually high-quality images \citep{koetzier2023deep}. 
For tubular structures in particular, voxel-wise overlap alone does not fully characterize topological correctness, motivating connectivity-aware measures such as clDice \citep{shit2021cldice}. A reconstructed CT image can therefore appear locally plausible while still containing broken, missing, or incorrectly connected vessels. These challenges motivate explicitly incorporating volumetric vascular structure into the reconstruction process, rather than relying on image appearance alone.


To address these limitations, we propose \textbf{VascuSR}, an iterative
anatomy-aware diffusion framework for CT SR that targets both image fidelity
and vascular anatomical plausibility. Rather than relying only on individual
2D slices, VascuSR represents vascular anatomy explicitly in 3D. This
representation provides richer information about vascular topology while
separating vascular structure from CT intensity and organ texture, making
anatomical refinement a simpler learning task. We therefore use a conditional
3D rectified flow to refine the vascular representation. A lightweight
multi-scale anatomy controller then transforms the refined representation into
conditioning features for the diffusion decoder. At each round, the vascular representation is re-extracted and refined
from the latest SR volume to guide the next reconstruction, so that
the anatomical condition and SR output are updated in turn through
the feedback loop.

\begin{samepage}
Our contributions are threefold:
\begin{itemize}
    \item We introduce explicit 3D vascular representations as
    anatomical guidance for slice-wise CT super-resolution.
    Separating vascular refinement from image reconstruction
    simplifies structural learning while retaining HR supervision
    for image fidelity.

    \item We develop an iterative refinement--conditioning mechanism that
    combines a conditional 3D rectified-flow vascular refiner with a lightweight
    multi-scale anatomy controller, allowing anatomical guidance
    to adapt to each updated SR volume.

    \item We evaluate VascuSR on two public vascular CT datasets across image quality, vascular overlap, and three-dimensional vascular structure. VascuSR reduces class-averaged component counts by over 20\% on both datasets and class-averaged MCD by over 10\% on MSD, while comparisons with direct joint optimization support the benefits of the decoupled design.

\end{itemize}
\end{samepage}

\begin{figure}[t!]
    \centering
    \includegraphics[width=\linewidth]{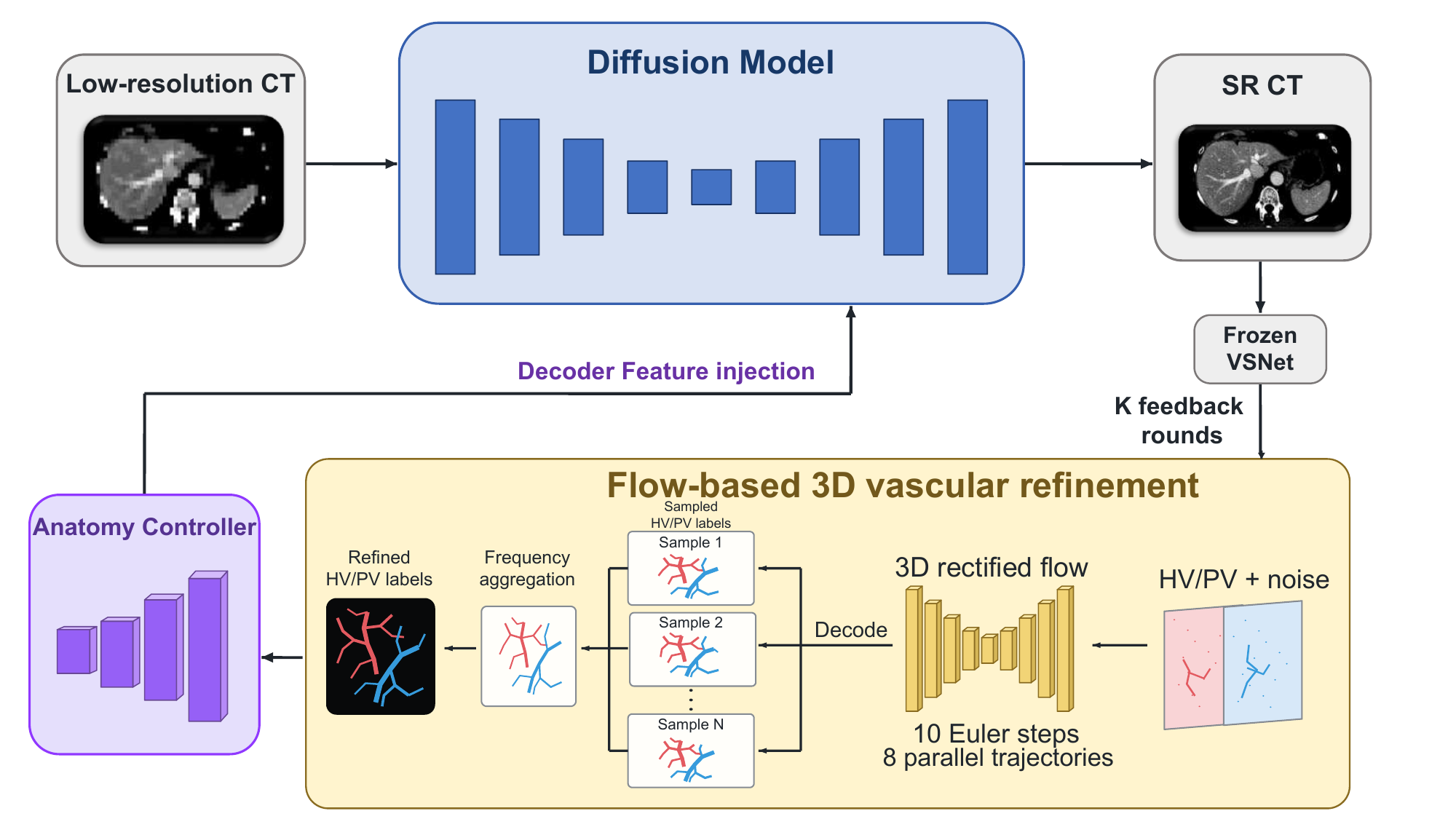}
    \vspace{-1.5em}
    \caption{\textbf{Overview of VascuSR.} Starting from an initial SR estimate, the recovered
    HV/PV anatomy is refined in three dimensions and fed back through a lightweight
    anatomy controller to condition the next diffusion-based reconstruction.
    Repeating this process for $K$ feedback rounds progressively couples image
    super-resolution with volumetric vascular refinement.}
    \label{fig:vascusr-overview}
\end{figure}

\vspace{-0.5em}
\section{Related Work}
\vspace{-0.2em}
\subsection{Super-Resolution and Generative Inverse Problems}
\vspace{-0.2em}




Super-resolution can be formulated as an ill-posed inverse problem, where a
high-resolution signal must be recovered from a degraded low-resolution
observation~\citep{wang2021deep}. Because multiple HR solutions may be consistent with the same LR
input, successful reconstruction requires appropriate priors over the missing
fine-scale structure. Deterministic SR methods learn such priors implicitly from paired
training data through a direct LR-to-HR mapping, using convolutional
architectures such as SRCNN, EDSR, and RCAN
\citep{dong2014learning,lim2017enhanced,zhang2018image},
or Transformer-based models such as SwinIR and HAT
\citep{liang2021swinir,chen2023hat}. These approaches achieve strong
distortion-oriented performance, but training with pixel-wise objectives can
produce overly smooth reconstructions.

Generative methods instead model a distribution over plausible HR solutions.
Adversarial approaches such as SRGAN use perceptual and adversarial objectives
to recover sharper image detail \citep{ledig2017photo}, while diffusion models
provide a probabilistic reconstruction framework through iterative denoising.
SR3 conditions the reverse diffusion process on the LR observation, whereas
ResShift constructs a residual-shifting trajectory between LR and HR
representations for more efficient sampling
\citep{saharia2023image,yue2023resshift}. StableSR and DiffBIR further exploit
pretrained latent diffusion models as generative priors for image restoration
\citep{wang2024stablesr,lin2024diffbir}. More broadly, this coarse-to-fine
inverse formulation is common across AI for Science, including medical imaging
and physical-system reconstruction, where generative priors are often combined
with domain-specific constraints to restrict the space of plausible solutions
\citep{chung2023dps,huang2024diffusionpde, yao2026posteriorbench}.

\subsection{Anatomy- and Task-Aware Medical Super-Resolution}



Medical SR methods incorporate domain knowledge primarily through either
observation consistency or anatomy- and task-specific priors. One line of work enforces consistency between the reconstruction
and the LR input in the image or measurement domain. \citet{chung2023mr}
reuse a score model trained for MR denoising for post-hoc
SR, alternating reverse-diffusion steps with image-level data-consistency
updates. GAN-CIRCLE regularizes bidirectional LR-HR mappings through cycle
consistency \citep{you2020gancircle}, CMISR iteratively corrects
re-reconstruction errors \citep{li2024cmisr}, and DynamicDPS applies adaptive
diffusion posterior guidance to low-field MRI enhancement
\citep{kim2025dynamicdps}. Such constraints improve agreement with the observed image or measurement.
However, this agreement alone does not ensure faithful recovery of fine
anatomical structures that are poorly resolved in the input.

A complementary line incorporates structural or task-specific information.
DiffMSR performs diffusion in a compact latent space to generate a
high-frequency-detail prior for multi-contrast MRI SR
\citep{li2024rethinking}, while DSDM combines image and structure streams with
Hessian-derived vessel- and blob-sensitive priors for CT SR
\citep{chu2023topology}. The dual-guidance SR network of \citet{chi2023dual} combines
segmentation-derived ROI guidance with inter-slice information
for CT super-resolution.
However, an important gap remains in how structural guidance can improve
vascular anatomical fidelity, and how this fidelity should be assessed
beyond conventional image-quality metrics.
\vspace{-3mm}


\section{Method}
\vspace{-0.3em}




\subsection{Problem Setup and Framework Overview}
\vspace{-0.3em}
\label{sec:problem-setup}

Given a low-resolution CT volume $x$, our goal is to reconstruct
a high-resolution volume $\widehat{y}$ that preserves both image
fidelity and vascular anatomical plausibility.
As illustrated in Figure~\ref{fig:vascusr-overview}, VascuSR combines
a slice-wise diffusion SR backbone, a 3D vascular refiner, and
an anatomy controller.
The backbone reconstructs high-resolution CT images, the refiner improves 3D vascular structure, and the controller injects the refined anatomy into the diffusion decoder.
The backbone and controller are trained jointly with HR supervision,
while the refiner learns vascular correction separately in the 3D vascular representation space, decoupling structural refinement from CT intensity reconstruction.
During inference, the vascular representation extracted from the
current SR volume is refined and passed through the controller
to guide the next reconstruction, forming an iterative feedback loop in which the anatomical condition is updated after each SR pass.
Appendix~\ref{app:finetuning-comparison} compares this decoupled formulation with directly optimizing image and anatomical objectives in the SR generator.

\vspace{-0.3em}
\subsection{Iterative Anatomy-Aware Diffusion}
\vspace{-0.3em}

Let $K\in\mathbb{N}$ denote the total number of feedback updates, and let
$r\in\{0,\ldots,K\}$ index the current state of the outer feedback loop.
We write
$x\in\mathbb{R}^{(X/s)\times(Y/s)\times Z}$ for the LR input CT volume,
where $s$ is the in-plane super-resolution scale factor, and
$\widehat y^{(r)}\in\mathbb{R}^{X\times Y\times Z}$ for the SR volume
after $r$ feedback updates. Here, $X$, $Y$, and $Z$ are the spatial
dimensions of the HR volume. Thus, $\widehat y^{(0)}$ is the initial SR
estimate, and $\widehat y^{(K)}$ is the final output $\widehat y$
introduced in Section~\ref{sec:problem-setup}.

Let $\theta$, $\phi$, and $\psi$ denote the parameters of the
SR backbone, anatomy controller $A_\phi$, and vascular refiner
$R_\psi$, respectively.
The backbone and controller together define the SR model
$p_{\theta,\phi}$.
We generate the initial SR volume $\widehat y^{(0)}$ conditioned on an all-zero 
vascular mask
$\mathbf{0}\in\{0,1\}^{2\times X\times Y\times Z}$, where the two channels correspond to hepatic veins (HV) and portal veins (PV):
\begin{equation}
    \widehat y^{(0)}
    \sim p_{\theta,\phi}\!\left(
        \,\cdot\mid x,A_\phi(\mathbf{0})
    \right).
    \label{eq:initial-sr}
\end{equation}

At each update $r\in\{0,\ldots,K-1\}$, the frozen VSNet
segmenter $S$~\citep{xu2025vsnet} produces vessel probabilities
$Q^{(r)}$, which the refiner $R_\psi$ converts into a binary
vascular condition $M^{(r)}$:
\begin{equation}
    \begin{aligned}
        Q^{(r)}
        &= S\!\left(\widehat y^{(r)}\right)
        \in [0,1]^{3\times X\times Y\times Z},\\
        M^{(r)}
        &= R_\psi\!\left(Q^{(r)}\right)
        \in \{0,1\}^{2\times X\times Y\times Z}.
    \end{aligned}
    \label{eq:vessel-parsing}
\end{equation}
In our hepatic-vessel experiments, the channels of $Q^{(r)}$
represent background, hepatic vein (HV), and portal vein (PV),
while $M^{(r)}$ contains the refined HV and PV masks.

The next SR volume is generated conditioned on $M^{(r)}$:
\begin{equation}
    \widehat y^{(r+1)}
    \sim p_{\theta,\phi}\!\left(
        \,\cdot\mid x,A_\phi\!\left(M^{(r)}\right)
    \right).
    \label{eq:iterative-generation}
\end{equation}
The LR input $x$ remains fixed throughout the loop, and each
update follows
\begin{equation*}
    \widehat y^{(r)}
    \xrightarrow{\,S\,}
    Q^{(r)}
    \xrightarrow{\,R_\psi\,}
    M^{(r)}
    \longrightarrow
    \widehat y^{(r+1)}.
\end{equation*}
We use K = 3 for MSD and K = 2 for HVM. The sensitivity to the number of feedback rounds is examined in
Appendix~\ref{app:iterative-feedback-ablation}.

\begin{table}[!t]
    \caption{Quantitative results for $\times 8$ CT super-resolution on
    (a) the MSD Task08
    HepaticVessel test set and (b) the HVM test set. In this table, VascuSR uses $K=3$
    on MSD and $K=2$ on HVM.
    The best and second-best results are shown in bold and with
    dashed underlining, respectively.}
    \label{tab:main-results}
    \centering
    \normalsize
    \setlength{\tabcolsep}{3pt}
    \begin{tabular}{lccccccc}
        \toprule
        Method & \multicolumn{2}{c}{Dice $\uparrow$}
        & \multicolumn{2}{c}{clDice $\uparrow$}
        & \multicolumn{3}{c}{Image quality} \\
        \cmidrule(lr){2-3}\cmidrule(lr){4-5}\cmidrule(lr){6-8}
        & HV & PV & HV & PV
        & PSNR $\uparrow$ & SSIM $\uparrow$ & LPIPS $\downarrow$ \\
        \midrule
        \multicolumn{8}{l}{\textbf{(a) Re-annotated MSD Task08 HepaticVessel}} \\
        Bicubic
            & .2618 & .2264 & .2475 & .1956 & 25.31 & .7742 & .3872 \\
        RCAN
            & .5465 & .5021 & .5616 & .4274
            & \dashuline{34.25} & .9124 & .1317 \\
        SwinIR
            & .5780 & .5623 & .5862 & .4626
            & \textbf{34.44} & \textbf{.9164} & .1262 \\
        HAT
            & .5720 & .5615 & .5782 & .4551
            & 34.21 & \dashuline{.9149} & .1270 \\
        ResShift
            & .6526 & .5923 & .5877 & .4460
            & 32.83 & .8896 & \dashuline{.0660} \\
        ResShift+Dice/clDice
            & \dashuline{.7010} & \dashuline{.6042}
            & \dashuline{.6385} & \dashuline{.4734}
            & 32.62 & .8823 & .0730 \\
        VascuSR (ours, $K{=}3$)
            & \textbf{.7072} & \textbf{.6404}
            & \textbf{.6446} & \textbf{.5286}
            & 32.77 & .8876 & \textbf{.0649} \\
        \midrule
        HR reference 
            & .7936 & .7610 & .7615 & .7352 & -- & -- & -- \\
        \midrule
        \multicolumn{8}{l}{\textbf{(b) HVM test set}} \\
        Bicubic
            & .4888 & .4962 & .4696 & .3851 & 19.24 & .6284 & .5043 \\
        RCAN
            & .5793 & .6751 & .5618 & .5530
            & 24.58 & .8379 & .1935 \\
        SwinIR
            & .5908 & .6804 & .5740 & \dashuline{.5588}
            & \textbf{24.76} & \textbf{.8410} & .1875 \\
        HAT
            & .5802 & .6760 & .5507 & .5442
            & \dashuline{24.60} & \dashuline{.8401} & .1829 \\
        ResShift
            & .5977 & .6914
            & .5687 & .5302
            & 23.50 & .7001 & \dashuline{.0743} \\
        ResShift+Dice/clDice
            & \dashuline{.6104} & \dashuline{.7001}
            & \dashuline{.5842} & .5469
            & 23.50 & .6983 & \textbf{.0740} \\
        VascuSR (ours, $K{=}2$)
            & \textbf{.6206} & \textbf{.7117}
            & \textbf{.6033} & \textbf{.5731}
            & 23.40 & .6990 & .0744 \\
        \midrule
        HR reference 
            & .7177 & .7917 & .7122 & .7149 & -- & -- & -- \\
        \bottomrule
    \end{tabular}
\end{table}

\subsection{Flow-Based 3D Vascular Refinement}
\label{sec:3D Vascular Refinement}
The vascular refiner $R_\psi$ learns to correct imperfect vessel predictions
using conditional 3D rectified flow~\citep{liu2023flow}, supervised by manual
HV/PV annotations. The full three-class probability volume $Q^{(r)}$
conditions refinement, while its HV and PV channels initialize the flow state.
Given the paired manual label
$M_{\mathrm{GT}}\in\{0,1\}^{2\times X\times Y\times Z}$, we define the
target and starting states as
\begin{equation}
    z_0 = 2M_{\mathrm{GT}}-1,
    \qquad
    z_1^{(r)} =
    2[Q_{\mathrm{HV}}^{(r)},Q_{\mathrm{PV}}^{(r)}]-1+\sigma\epsilon,
    \quad \epsilon\sim\mathcal{N}(0,I).
    \label{eq:segflow-endpoints}
\end{equation}
Let $\tau\in[0,1]$ denote the flow time, where $\tau=1$ corresponds to
the noisy initial state $z_1^{(r)}$ and $\tau=0$ corresponds to the
target state $z_0$. The parameter $\sigma$ controls the noise amplitude.
The mapping $a\mapsto 2a-1$ encodes binary labels from $\{0,1\}$ to
$\{-1,+1\}$, allowing foreground to be decoded using a zero threshold.

During training, we sample the flow time $\tau\sim\mathcal U(0,1)$ and
construct $z_\tau^{(r)}=(1-\tau)z_0+\tau z_1^{(r)}$. A 3D U-Net
$v_\psi$, with parameters $\psi$, predicts the velocity from this state, flow
time $\tau$, and condition $Q^{(r)}$. We train it to match the velocity
$z_1^{(r)}-z_0$ along the linear path:
\begin{equation}
    \mathcal{L}_{\mathrm{flow}}
    =\mathbb{E}
    \left[
        \left\|v_\psi(z_\tau^{(r)},\tau\mid Q^{(r)})
        -\left(z_1^{(r)}-z_0\right)\right\|_2^2
    \right].
    \label{eq:segflow-loss}
\end{equation}
The expectation is over paired training examples, flow time, and Gaussian
noise.

\begin{samepage}
At inference, we run $N=8$ trajectories in parallel with independent initial
noise and aggregate their decoded labels.
Each trajectory is integrated from $\tau=1$ to $\tau=0$ using ten Euler
steps, with $Q^{(r)}$ fixed throughout. Let $\widehat z_0^{(r,n)}$ denote the
predicted endpoint of trajectory $n\in\{1,\ldots,N\}$. The deterministic
decoder $\mathcal D$ applies a zero foreground threshold followed by HV/PV
channel competition. For voxel $j$ and class $c\in\{\mathrm{HV},\mathrm{PV}\}$,
the frequency of decoded class labels is
\begin{equation}
    F_c^{(r)}(j)
    = \frac{1}{N}\sum_{n=1}^{N}
      \mathbf{1}\!\left[
        \mathcal{D}\!\left(\widehat{z}_0^{(r,n)}\right)(j)=c
      \right].
    \label{eq:segflow-ensemble}
\end{equation}
Applying a validation-selected frequency threshold and HV/PV class competition
to $F^{(r)}$ yields $M^{(r)}$, the anatomical condition for the next SR round.
\end{samepage}
\subsection{Multi-Scale Anatomy Controller}
\label{sec:anatomy-controller}

The vascular refiner in Section~\ref{sec:3D Vascular Refinement} produces the
refined 3D vascular condition $M^{(r)}$, while the anatomy controller
transfers this condition to the slice-wise SR backbone.

To guide slice-wise SR with the refined 3D anatomy, a lightweight
convolutional controller takes the corresponding axial mask slices from
$M^{(r)}$ as input. These slices form a batch
$m^{(r)}\in\{0,1\}^{B\times2\times H\times W}$, where $B$ is the batch
size, the two channels represent HV and PV, and $H\times W$ is the slice
size. Let $l\in\{1,\ldots,L\}$ index the decoder injection levels and $t$
the SR diffusion step. During the next reconstruction
$\widehat y^{(r+1)}$, the controller produces an anatomical control feature
at each level:
\begin{equation}
    c_l^{(r,t)}
    =
    A_{\phi,l}\!\left(m^{(r)},t\right)
    \in
    \mathbb{R}^{B\times D_l\times H_l\times W_l}.
    \label{eq:control-feature}
\end{equation}
Here, $A_{\phi,l}$ denotes the controller mapping at level $l$.
Let $h_l^{(r+1,t)}$ denote the corresponding SR decoder feature.
We add the anatomical control feature to obtain the conditioned
decoder feature:
\begin{equation}
    \widehat h_l^{(r+1,t)}
    = h_l^{(r+1,t)} + c_l^{(r,t)}.
    \label{eq:decoder-injection}
\end{equation}
Hyperparameter settings and training and inference details
for the architecture described above are provided in
Appendix~\ref{app:refinement-details}.

\begingroup
\raggedbottom
\let\clearpage\relax
\section{Experiments and Results}

\subsection{Datasets}

We use two publicly available contrast-enhanced CT datasets for $\times8$ SR,
pairing each $512\times512$ HR slice with a $64\times64$ LR input. All methods
use identical preprocessing within each dataset, and all splits and analyses
are patient-level to avoid leakage between correlated slices.
No patient in the reported test sets is used to train the SR models, VSNet, or
FlowSeg3D (the vascular refiner). Patient splits and additional implementation details are provided
in Appendix~\ref{app:implementation-details}.

\textbf{Re-annotated MSD Task08 HepaticVessel.}
The official Medical Segmentation Decathlon (MSD) Task08 dataset comprises 443
CT volumes: 303 labeled training volumes and 140 challenge-test volumes whose
annotations are not publicly released~\citep{antonelli2022msd}.
VSNet re-annotates the 303 training volumes with separate HV and PV masks.

\textbf{HVM.}
The public Hepatic Vessel Map (HVM) dataset~\citep{xie2026hvm} contains 282
dual-center portal-venous-phase contrast-enhanced CT volumes with separate HV
and PV annotations. We exclude one case with an incomplete image--label pair,
leaving 281 complete CT--HV--PV triplets. Because the released volumes fall into three stored-intensity domains
and are not uniformly raw-HU CT, we clip all volumes to their shared range
$[-40,160]$ before normalization.

\subsection{Baselines}
We compare VascuSR with four groups of baselines:
\begin{itemize}
    \item \textbf{Interpolation.}
    Bicubic interpolation provides a non-learned reference, reconstructing
    HR images without learned image or anatomical priors.

    \item \textbf{Deterministic SR.}
    RCAN~\citep{zhang2018image}, SwinIR~\citep{liang2021swinir}, and
    HAT~\citep{chen2023hat} represent convolutional and Transformer-based
    reconstruction methods. 

    \item \textbf{Diffusion SR.}
    ResShift~\citep{yue2023resshift} performs SR through
    residual-shifting diffusion and serves as the backbone of VascuSR.

    \item \textbf{Joint image and anatomical optimization.}
    Extends ResShift by adding Dice and soft-clDice losses,
    computed through a frozen VSNet segmenter, to its image
    reconstruction objective during fine-tuning.
    Details are provided in
    Appendix~\ref{app:finetuning-comparison}.
\end{itemize}

Learned baselines are trained or fine-tuned on the same LR--HR
training pairs within each dataset. We use each method's official
training recipe as a starting point and select hyperparameters
using the validation set.



\subsection{Evaluation Metrics}
\label{sec:evaluation-metrics}

\paragraph{Image reconstruction fidelity.}
We report peak signal-to-noise ratio (PSNR), structural similarity
index (SSIM)~\citep{wang2004image}, and learned perceptual image
patch similarity (LPIPS)~\citep{zhang2018perceptual}.
PSNR quantifies pixel-level fidelity based on mean squared error,
SSIM compares local luminance, contrast, and structure, and LPIPS
measures perceptual distance in a learned feature space.

\paragraph{Vessel segmentation agreement.}
We apply the same fixed VSNet segmenter to every reconstructed volume and
compare its HV and PV predictions with the reference annotations. Dice measures volumetric overlap, while clDice
measures skeleton-based agreement~\citep{shit2021cldice}.

\paragraph{Three-dimensional vascular structure.}
To complement Dice and clDice, we further assess vascular
fragmentation and the spatial alignment of vessel centerlines
with the reference anatomy.

We use 26-connectivity to compute connected components separately for HV and
PV. For $c\in\{\mathrm{HV},\mathrm{PV}\}$, \#CC counts components, and the
largest-component fraction is $|C_{\max,c}|/|P_c|$, where $P_c$ is the
predicted foreground of class $c$ and $C_{\max,c}$ its largest connected
component.

Mean centerline distance (MCD) measures the geometric discrepancy between
predicted and reference vessel trajectories~\citep{zhang2025rapid}. Let $C_P$
and $C_G$ denote centerline point sets extracted from the predicted and manual
reference masks, respectively, in physical coordinates (mm). We compute
\begin{equation}
    \operatorname{MCD}(C_P,C_G)
    = \frac{1}{2}\left[
        \frac{1}{|C_P|}\sum_{p\in C_P}\min_{g\in C_G}\|p-g\|_2
        + \frac{1}{|C_G|}\sum_{g\in C_G}\min_{p\in C_P}\|g-p\|_2
      \right],
    \label{eq:mcd}
\end{equation}
where lower values indicate closer centerline agreement.

\subsection{Comparison with Super-Resolution Baselines}

On the re-annotated MSD Task08 set, SwinIR achieves the highest PSNR and SSIM,
consistent with the emphasis of regression-based SR on pixel-level fidelity.
Nevertheless, VascuSR achieves the best LPIPS, which captures perceptual
similarity in texture and local structure, and leads all methods across the
Dice and clDice measures of anatomical fidelity. These results indicate that VascuSR focuses more on improving vascular
structure recovery in SR images while preserving the backbone's
image reconstruction capability.
A detailed comparative analysis of VascuSR and ResShift+Dice/clDice across
local appearance, vascular topology, and other aspects is provided in
Appendix~\ref{app:finetuning-comparison}.
Additional results on vessel continuity and centerline accuracy
are reported in Table~\ref{tab:vascular-structure}.
 
\subsection{Three-Dimensional Vascular Structure}
\label{sec:vascular-structure}

\begin{table}[!t]
    \caption{Three-dimensional vascular structure on (a) MSD Task08 HepaticVessel test set
    and (b) HVM test set. Bold marks the best SR result in each column.}
    \label{tab:vascular-structure}
    \label{tab:cc-continuity}
    \label{tab:msd-mcd}
    \label{tab:hvm-vascular-structure}
    \label{tab:hvm-cc-continuity}
    \label{tab:hvm-mcd}
    \centering
    \normalsize
    \setlength{\tabcolsep}{4pt}
    \begin{tabular}{lcccccc}
        \toprule
        & \multicolumn{3}{c}{HV} & \multicolumn{3}{c}{PV} \\
        \cmidrule(lr){2-4}\cmidrule(lr){5-7}
        Method & \#CC $\downarrow$ & \makecell{Largest-CC\\frac $\uparrow$} & \makecell{MCD\\(mm) $\downarrow$}
               & \#CC $\downarrow$ & \makecell{Largest-CC\\frac $\uparrow$} & \makecell{MCD\\(mm) $\downarrow$} \\
        \midrule
        \multicolumn{7}{c}{\textbf{(a) MSD Task08 HepaticVessel}} \\
        \midrule
        Bicubic & 27.6 & .5178 & 18.128 & \textbf{13.0} & .6432 & 18.171 \\
        RCAN & 31.1 & .7773 & 6.753 & 29.0 & .6801 & 8.700 \\
        SwinIR & 31.6 & .8377 & 6.499 & 28.1 & .7874 & 8.632\\
        HAT & 33.1 & .8125 & 6.517 & 29.0 & .7852 & 8.453 \\
        ResShift & 31.0 & .9077 & 6.046 & 26.0 & .8843 & 8.088 \\
        ResShift+Dice/clDice & 39.6 & .8936 & 5.652 & 30.4 & .8249 & 7.537 \\
        VascuSR (ours, $K{=}3$)
            & \textbf{22.7} & \textbf{.9382} & \textbf{5.231}
            & 22.4 & \textbf{.8931} & \textbf{6.635} \\
                   \midrule
        Manual GT & 1.0 & 1.0000 & 0.0000 & 1.0 & 1.0000 & 0.0000 \\
        \midrule
        \multicolumn{7}{c}{\textbf{(b) HVM}} \\
        \midrule
        Bicubic & \textbf{28.9} & .493 & 7.507 & \textbf{23.3} & .754 & 7.876 \\
        RCAN & 52.8 & .456 & 5.732 & 69.8 & .821 & 5.539 \\
        SwinIR & 51.5 & .467 & 5.673 & 65.2 & .843 & 5.264 \\
        HAT & 76.2 & .460 & 5.651 & 96.8 & .812 & 5.718 \\
        ResShift & 48.7 & .497 & 5.272 & 66.7 & .874 & 5.004 \\
        ResShift+Dice/clDice & 53.3 & .494 & 5.233 & 62.8 & .882 & 4.951 \\
        VascuSR (ours, $K{=}2$)
            & 31.8 & \textbf{.524} & \textbf{5.029}
            & 45.0 & \textbf{.923} & \textbf{4.743} \\
        \midrule
        Manual GT & 7.3 & .601 & 0.0000 & 7.8 & .996 & 0.0000 \\
        \bottomrule
    \end{tabular}
\end{table}

Across MSD and HVM (Table~\ref{tab:vascular-structure}), VascuSR reduces component counts
and increases largest-component fractions for both classes relative to
ResShift. Bicubic has the lowest PV count, but its lower largest-component
fraction illustrates why a low count alone does not imply better continuity:
under-segmentation can also reduce counts.
On MSD, PV MCD decreases from 8.0877\,mm with ResShift to 6.6351\,mm, a reduction of
18.0\%. These results show that VascuSR reconstructs less fragmented
vascular structures with centerlines that more closely follow
the reference anatomy.

Manual inspection of HVM annotations reveals isolated fragments and incomplete
HV coverage at the hepatocaval junctions, where hepatic veins drain into the
inferior vena cava. The resulting fragments and separate major venous subtrees
explain the elevated component counts and HV largest-component fraction of
approximately 0.60.

An ablation study of anatomical conditioning and iterative
feedback is provided in
Appendix~\ref{app:iterative-feedback-ablation}. It shows that VascuSR benefits from anatomical guidance and that
Dice and clDice improve with feedback, largely stabilizing after
two to three rounds.


\subsection{Qualitative Analysis}

We further provide qualitative visualizations to examine local vascular
changes that are not fully captured by aggregate scores. Modest changes in
Dice can conceal clear local structural differences. In
Figure~\ref{fig:qualitative-comparison}, the two central PV portions appear
separated in the ResShift reconstruction. After three feedback rounds,
VascuSR improves their visible connection and recovers additional vessel
contours visible in the HR reference.

Figure~\ref{fig:qualitative-3d} further shows three-dimensional vascular
morphology. After three feedback rounds, the vascular trees are visibly more
continuous, while several small isolated components present in the initial
SR-derived segmentation are removed. Additional qualitative results on MSD and HVM
are provided in Appendix~\ref{app:hvm-qualitative}.

\begin{figure}[!t]
    \centering
    \includegraphics[width=.9\linewidth]{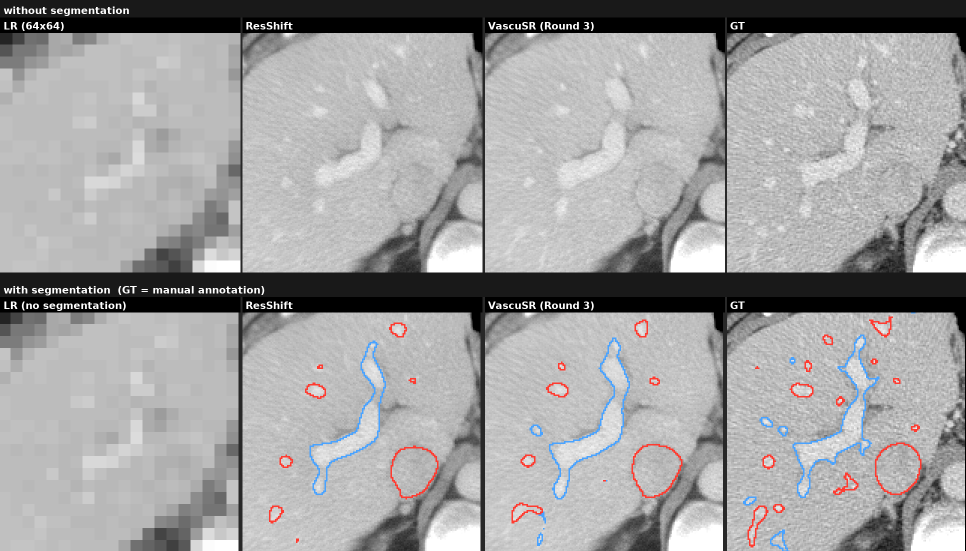}

    \caption{MSD case 416 at axial slice $z=094$. Columns: LR, ResShift,
    VascuSR (round 3), and HR. Top: image intensities; bottom: HV (red) and
    PV (blue) contour overlays.}
    \label{fig:qualitative-comparison}

    \vspace{2mm}
    \includegraphics[width=\linewidth]{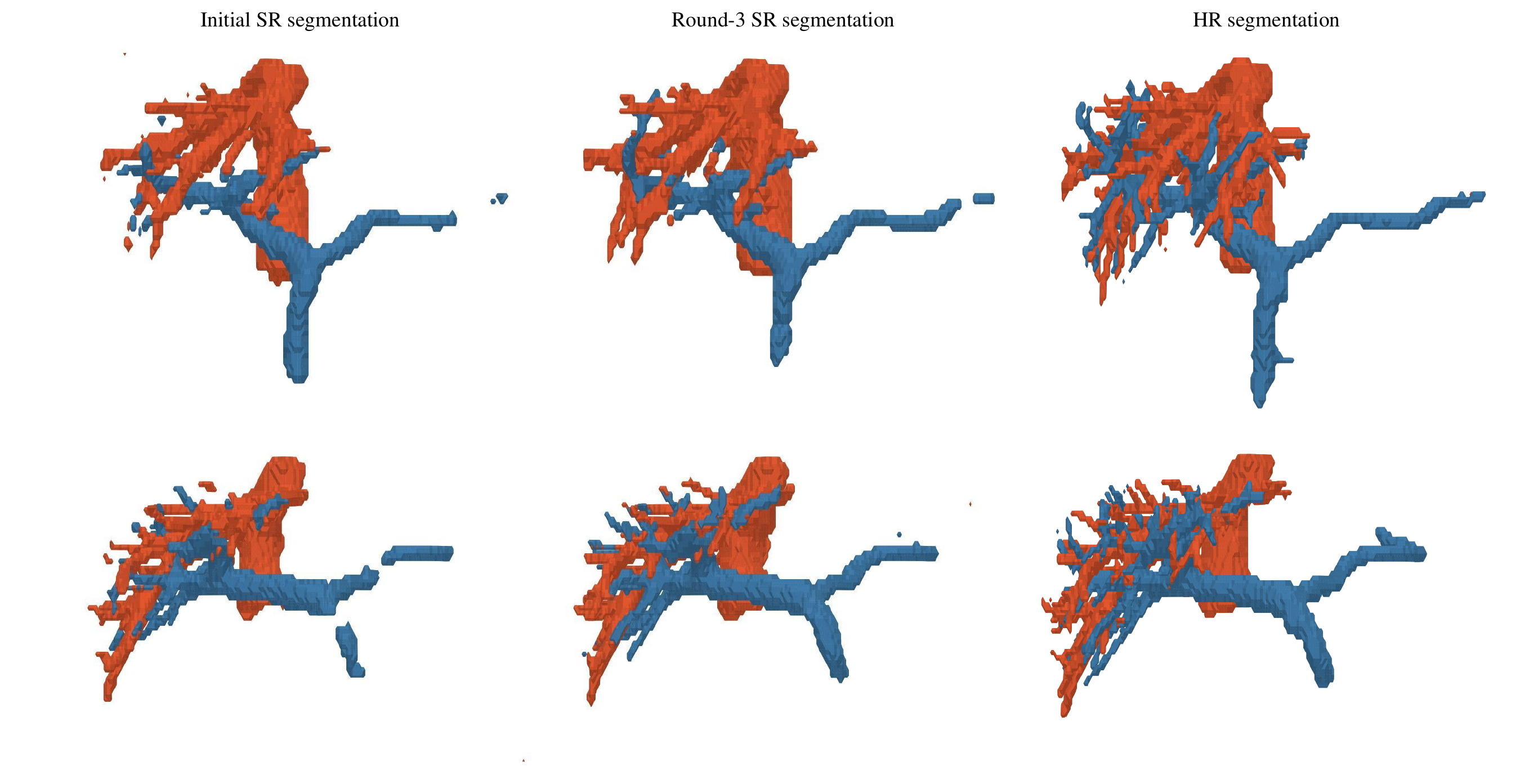}

    \caption{Three-dimensional vascular comparison on MSD cases 416 (top) and
    411 (bottom). Columns: initial SR (round 0), round-3 SR, and HR segmentations.
    Red: HV; blue: PV.}
    \label{fig:qualitative-3d}
\end{figure}

\newpage
\subsection{Parameter and Runtime Overhead}

We report the parameter counts of the components added to the pipeline and
their inference time. As shown in Table~\ref{tab:efficiency}, the anatomy
controller and FlowSeg3D introduce only modest parameter overhead relative to
ResShift. With eight refinement trajectories sampled in parallel, a
three-round configuration processes one volume in approximately 6 min, which
is acceptable for offline volumetric analysis.

\begin{table}[H]
    \caption{Parameter and runtime overhead. (a) The parameter total includes
    the ResShift backbone, anatomy controller, and FlowSeg3D refiner; GFLOPs
    are measured end to end for one SR pass. (b) Runtime per three-dimensional
    volume; FlowSeg3D uses $N=8$ parallel trajectories.}
    \label{tab:efficiency}
    \centering
    \normalsize
    \textbf{(a) Parameters and SR-generation compute}\\[1mm]
    \setlength{\tabcolsep}{3.5pt}
    \begin{tabular}{lccccc}
        \toprule
        Method & \makecell{Backbone incl.\\VQ-f4 (M)}
        & \makecell{Controller\\(M)} & \makecell{FlowSeg3D\\(M)}
        & \makecell{Total\\(M)} & \makecell{End-to-end GFLOPs/\\SR pass} \\
        \midrule
        ResShift & 172.01 & -- & -- & 172.01 & 6248.69 \\
        VascuSR & 172.01 & 1.44 & 16.25 & 189.70 & 6284.01 \\
        \bottomrule
    \end{tabular}

    \vspace{2mm}
    \textbf{(b) Runtime per volume (s)}\\[1mm]
    \setlength{\tabcolsep}{4pt}
    \begin{tabular}{lcccc}
        \toprule
        Pass & SR generation & VSNet & \makecell{FlowSeg3D\\($N=8$)} & Total \\
        \midrule
        Initial zero-mask & $18.41\pm1.83$ & -- & -- & $18.41\pm1.83$ \\
        Feedback round 1 & $18.40\pm1.82$ & $1.94\pm0.43$ & $102.22\pm17.87$ & $122.56\pm17.53$ \\
        Feedback round 2 & $18.44\pm1.85$ & $1.95\pm0.43$ & $97.59\pm13.06$ & $117.98\pm13.19$ \\
        Feedback round 3 & $18.41\pm1.83$ & $1.95\pm0.44$ & $100.36\pm17.78$ & $120.72\pm17.63$ \\
        \bottomrule
    \end{tabular}
\end{table}

\endgroup

\section{Conclusion and Discussion}

We introduced VascuSR, an iterative anatomy-aware diffusion framework for
vascular CT super-resolution that decouples three-dimensional vascular
refinement from CT intensity reconstruction and connects them through
anatomical conditioning. Across MSD Task08 and HVM,
VascuSR improves Dice and clDice over the matched ResShift backbone while
maintaining comparable image-level reconstruction quality. Connectivity and
centerline analyses further show reduced fragmentation and closer agreement
with reference vessel geometry. Further comparisons with joint optimization of image reconstruction
and anatomical objectives support the benefits of our decoupled
design in local image fidelity, vascular structure, and training
efficiency. These findings support learning an
anatomical prior in vascular label space and transferring it to SR through
conditioning, while retaining HR-supervised reconstruction as the image
objective.

\paragraph{Limitations and Future Work.}
The current framework performs SR slice-wise and relies on a fixed vessel segmenter and separately trained 3D refiner, which may limit volumetric consistency and allow errors to accumulate across feedback iterations. We also present exploratory results on coronary CTA in
Appendix~\ref{app:imagecas}, extending the evaluation beyond
hepatic vessels. However, larger-scale evaluation across diverse vascular
systems remains necessary. Future work will investigate fully volumetric and more tightly integrated reconstruction frameworks, adaptive or more efficient refinement, and broader evaluation across anatomies, acquisition settings, and downstream vascular-analysis tasks.



\subsection*{Acknowledgments}
This work is supported in part by the U.S. Army Research Office under Army-ECASE award W911NF-07-R-0003-03, the U.S. Department Of Energy, Office of Science, ARPA-H-SOL-24-101 program, IARPA HAYSTAC Program, NSF Grants \#2205093, \#2146343, \#2134274, \#2441832 and CDC-RFA-FT-23-0069.

\clearpage
\bibliographystyle{iclr2027_conference}
\bibliography{references}

\clearpage
\appendix
\raggedbottom
\section{Appendix}
\label{app:topology}

\subsection{Training and Inference Details}
\label{app:refinement-details}
\label{app:implementation-details}

\paragraph{Patient splits.}
Table~\ref{tab:patient-splits} reports the patient-level dataset partitions.
For MSD, 81 of the 243 training patients are used to train FlowSeg3D and the
remaining 162 to train the joint SR/controller model. For HVM, the corresponding
counts are 75 and 148 out of 223 training patients. The VSNet is trained on all patients of the training sets.

\begin{table}[H]
    \caption{Patient-level dataset partitions.}
    \label{tab:patient-splits}
    \centering
    \normalsize
    \setlength{\tabcolsep}{14pt}
    \begin{tabular}{lccc}
        \toprule
        Dataset & Training & Validation & Test \\
        \midrule
        Re-annotated MSD Task08 & 243 & 30 & 30 \\
        HVM & 223 & 29 & 29 \\
        \bottomrule
    \end{tabular}
\end{table}

\paragraph{Controller training conditions.}
During joint training of the SR backbone and anatomy controller,
we sample anatomical conditions from three sources: raw VSNet labels
from HR images, FlowSeg3D-refined labels from SR images, and an
all-zero mask. The corresponding sampling probabilities are
0.35, 0.35, and 0.30 on MSD, and
0.425, 0.425, and 0.15 on HVM, respectively.

\paragraph{SR training objective.}
The backbone and controller are jointly trained from scratch with the official ResShift objective. The VQ-f4 autoencoder
remains frozen, and the LPIPS reference is its round-trip reconstruction of
the HR target. 
\paragraph{SR optimization and inference.}
We use AdamW with a learning rate of $5\times10^{-5}$ and a global batch
size of 16, and use the checkpoint at 50,000 updates. Training and inference
use the four-step ResShift diffusion process. Training is performed with
DDP on four NVIDIA RTX PRO 6000 Blackwell GPUs.

\paragraph{FlowSeg3D inputs and training.}
On both datasets, the velocity network is a 3D U-Net with 24 base channels,
trained on $128\times128\times96$ patches with batch size 2 and an 80\%
probability of foreground-centered sampling. We use AdamW with learning rate
$2\times10^{-4}$, 500 warm-up iterations, and EMA decay $0.999$, optimizing
the velocity MSE in Equation~\ref{eq:segflow-loss}. The noise amplitude in Equation~\ref{eq:segflow-endpoints}
is $\sigma=1.0$ during both training and inference on MSD and HVM.

\paragraph{FlowSeg3D inference.}
The region of interest is the VSNet-predicted vessel bounding box expanded
by 24 voxels. Velocity predictions use sliding windows with 50\% overlap.
We integrate $N=8$ independent-noise trajectories in parallel from $\tau=1$
to $\tau=0$ with ten Euler steps, keeping the probability condition fixed.
After hard decoding, class frequencies are aggregated as in
Equation~\ref{eq:segflow-ensemble}. We retain frequencies strictly above $0.3$ on MSD and $0.4$ on HVM,
corresponding to at least three and four votes out of eight, respectively.
These thresholds are selected separately on the validation set
of each dataset. If both classes pass, the higher frequency
wins, with ties assigned to HV.

\clearpage
\subsection{Ablation Study: Iterative Anatomical Feedback}
\label{app:iterative-feedback-ablation}
\label{app:hvm-iterative-feedback}

We examine how anatomical fidelity and image quality change over successive
feedback rounds. On MSD (Table~\ref{tab:condition-ablation}), all five feedback
rounds improve the reported Dice and clDice scores over ResShift. These metrics
largely stabilize by round 3, with later rounds showing fluctuations rather
than consistent gains. Image-quality metrics
remain similar across feedback rounds.

\begin{table}[H]
    \caption{Effect of iterative anatomical feedback on the MSD Task08
    HepaticVessel test set. The 0-mask setting retains the controller architecture
    without anatomical conditioning.
    The best result in each column is shown in bold, including ties.}
    \label{tab:condition-ablation}
    \centering
    \normalsize
    \setlength{\tabcolsep}{4pt}
    \begin{tabular}{lccccccc}
        \toprule
        \multirow{2}{*}{Arm}
            & \multicolumn{2}{c}{Dice $\uparrow$}
            & \multicolumn{2}{c}{clDice $\uparrow$}
            & \multicolumn{3}{c}{Image quality} \\
        \cmidrule(lr){2-3}\cmidrule(lr){4-5}\cmidrule(lr){6-8}
            & HV & PV & HV & PV
            & PSNR $\uparrow$ & SSIM $\uparrow$ & LPIPS $\downarrow$ \\
        \midrule
        ResShift (baseline)
            & .6526 & .5923 & .5877 & .4460
            & \textbf{32.83} & \textbf{.8896} & .0660 \\
        0-mask
            & .6587 & .5644 & .5909 & .4286 & 32.78 & .8877 & .0650 \\
        VascuSR, round 1
            & \textbf{.7077} & .6334 & .6381 & .5158
            & 32.77 & .8876 & \textbf{.0649} \\
        VascuSR, round 2
            & .7050 & .6337 & .6431 & .5267
            & 32.77 & .8876 & \textbf{.0649} \\
        VascuSR, round 3
            & .7072 & \textbf{.6404} & .6446 & \textbf{.5286}
            & 32.77 & .8876 & \textbf{.0649} \\
        VascuSR, round 4
            & .6932 & .6365 & .6410 & .5203 & 32.77 & .8876 & .0650 \\
        VascuSR, round 5
            & .7039 & .6370 & \textbf{.6503} & .5235
            & 32.77 & .8876 & .0650 \\
        \bottomrule
    \end{tabular}
\end{table}

Table~\ref{tab:hvm-condition-ablation} reports the corresponding HVM
ablation. HVM reaches its strongest overall results around rounds 2--3.

\begin{table}[H]
    \caption{Effect of iterative anatomical feedback on HVM. The 0-mask
    setting (R0) retains the controller architecture without anatomical
    conditioning. The best result in each column is shown in bold.}
    \label{tab:hvm-condition-ablation}
    \centering
    \normalsize
    \setlength{\tabcolsep}{4pt}
    \begin{tabular}{lccccccc}
        \toprule
        \multirow{2}{*}{Arm}
            & \multicolumn{2}{c}{Dice $\uparrow$}
            & \multicolumn{2}{c}{clDice $\uparrow$}
            & \multicolumn{3}{c}{Image quality} \\
        \cmidrule(lr){2-3}\cmidrule(lr){4-5}\cmidrule(lr){6-8}
            & HV & PV & HV & PV
            & PSNR $\uparrow$ & SSIM $\uparrow$ & LPIPS $\downarrow$ \\
        \midrule
        ResShift (baseline)
            & .5977 & .6914 & .5687 & .5302
            & \textbf{23.50}
            & \textbf{.7001}
            & \textbf{.0743} \\
        0-mask (R0)
            & .5928 & .6888 & .5637 & .5289
            & 23.40 & .6990& .0744 \\
        VascuSR, round 1
            & .6191 & .7113 & .5980 & .5689
            & 23.41 & .6990 & .0744 \\
        VascuSR, round 2
            & \textbf{.6206} & \textbf{.7117}
            & \textbf{.6033} & .5731
            & 23.40 & .6990 & .0744 \\
        VascuSR, round 3
            & .6191 & .7111 & .6028 & \textbf{.5768}
            & 23.40 & .6990 & .0745 \\
        VascuSR, round 4
            & .6168 & .7088 & .6012 & .5722
            & 23.40 & .6990 & .0745 \\
        VascuSR, round 5
            & .6149 & .7069 & .5979 & .5724
            & 23.40 & .6990 & .0745 \\
        \bottomrule
    \end{tabular}
\end{table}

The 0-mask setting retains the additional controller parameters
but does not consistently improve Dice or clDice over ResShift.
This indicates that VascuSR's anatomical gains depend on anatomical
conditioning rather than increased model capacity alone.

\clearpage
\subsection{Additional Qualitative Results}
\label{app:qualitative-results}
\label{app:hvm-qualitative}

\begin{figure}[H]
    \centering
    \includegraphics[width=.94\linewidth]{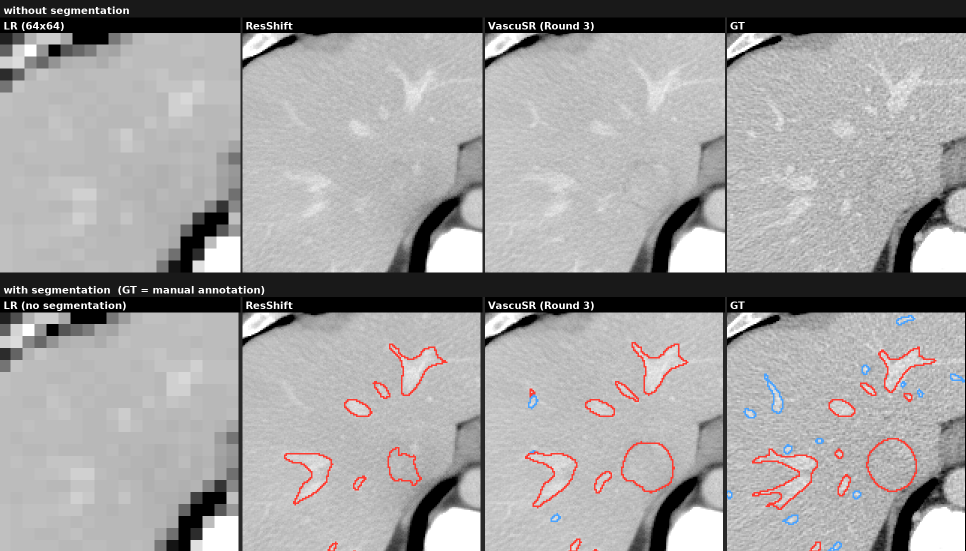}

    \caption{Additional qualitative comparison on held-out MSD case 416 at
    axial slice $z=114$. Columns show LR, ResShift, VascuSR after three
    feedback rounds, and the HR reference (GT). The upper row shows intensity
    crops; the lower row adds vascular contours to the reconstructed and HR
    images. Red and blue denote HV and PV, respectively.}
    \label{fig:qualitative-msd-extra}
\end{figure}

In Figure~\ref{fig:qualitative-msd-extra}, a low-contrast, approximately
circular HV cross-section in the lower-right region is only weakly and
irregularly delineated by ResShift. Anatomical feedback makes its boundary
more regular and the predicted contour closer to the manual annotation.

\clearpage
\begin{figure}[H]
    \centering

    \textbf{(a)}\\[-1mm]
    \par\vspace{2mm}
    \includegraphics[width=\linewidth]{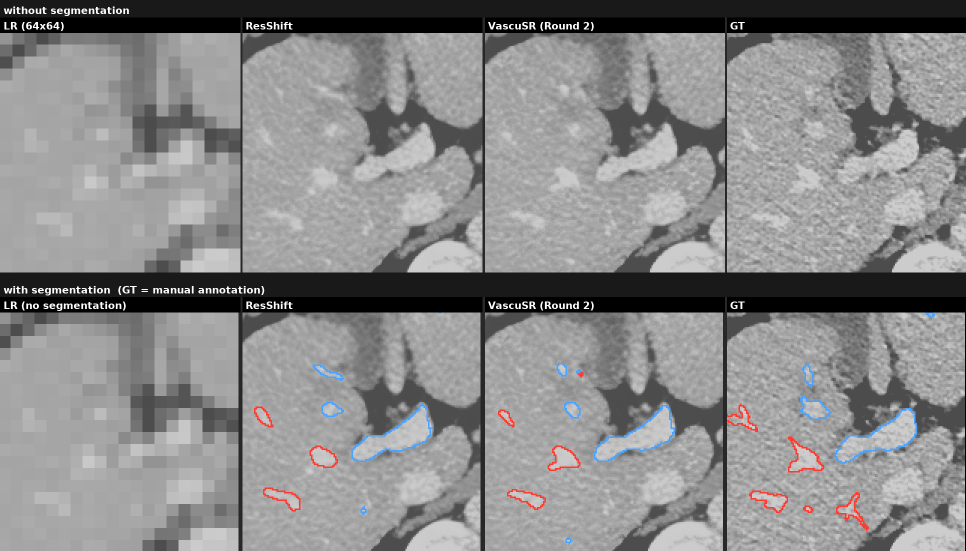}

    \vspace{0.5mm}

    \textbf{(b)}\\[-1mm]
    \par\vspace{2mm}
    \includegraphics[width=\linewidth]{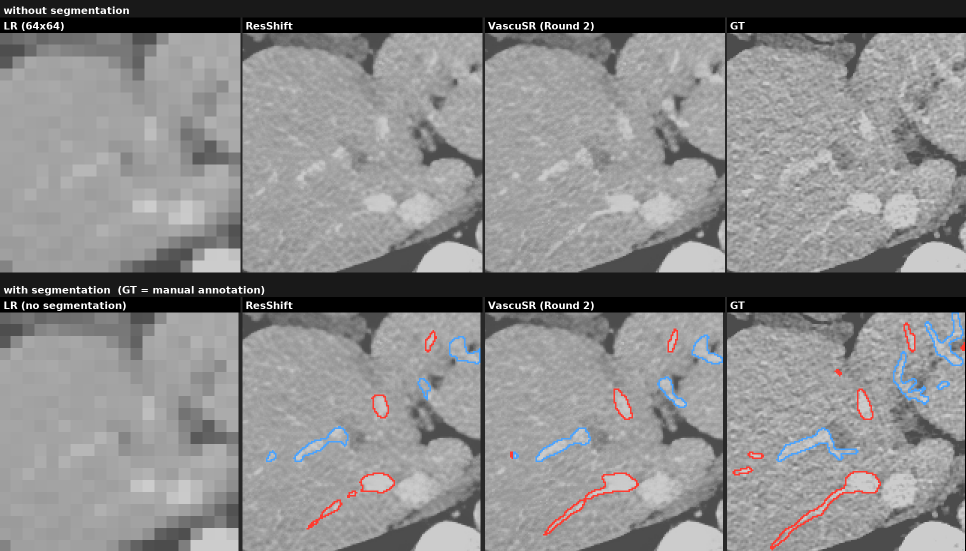}

    \caption{Additional qualitative comparison on held-out HVM case 109 at
    axial slices (a) $z=179$ and (b) $z=204$. Columns show LR, ResShift,
    VascuSR after two feedback rounds, and the HR reference (GT). The upper
    row of each panel shows intensity crops; the lower row adds vascular
    contours to the reconstructed and HR images. Red and blue denote HV
    and PV, respectively; HR contours are manual annotations.}
    \label{fig:hvm-qualitative-2d}
\end{figure}

In panel (a), the central HV cross-section appears rounded in ResShift,
whereas VascuSR recovers a more angular shape that more closely resembles
the HR reference. In panel (b), the elongated HV and PV structures near
the center appear interrupted in ResShift and become more continuous
with VascuSR.

\clearpage
\begin{figure}[H]
    \centering
    \includegraphics[
        width=\linewidth,
        height=.86\textheight,
        keepaspectratio
    ]{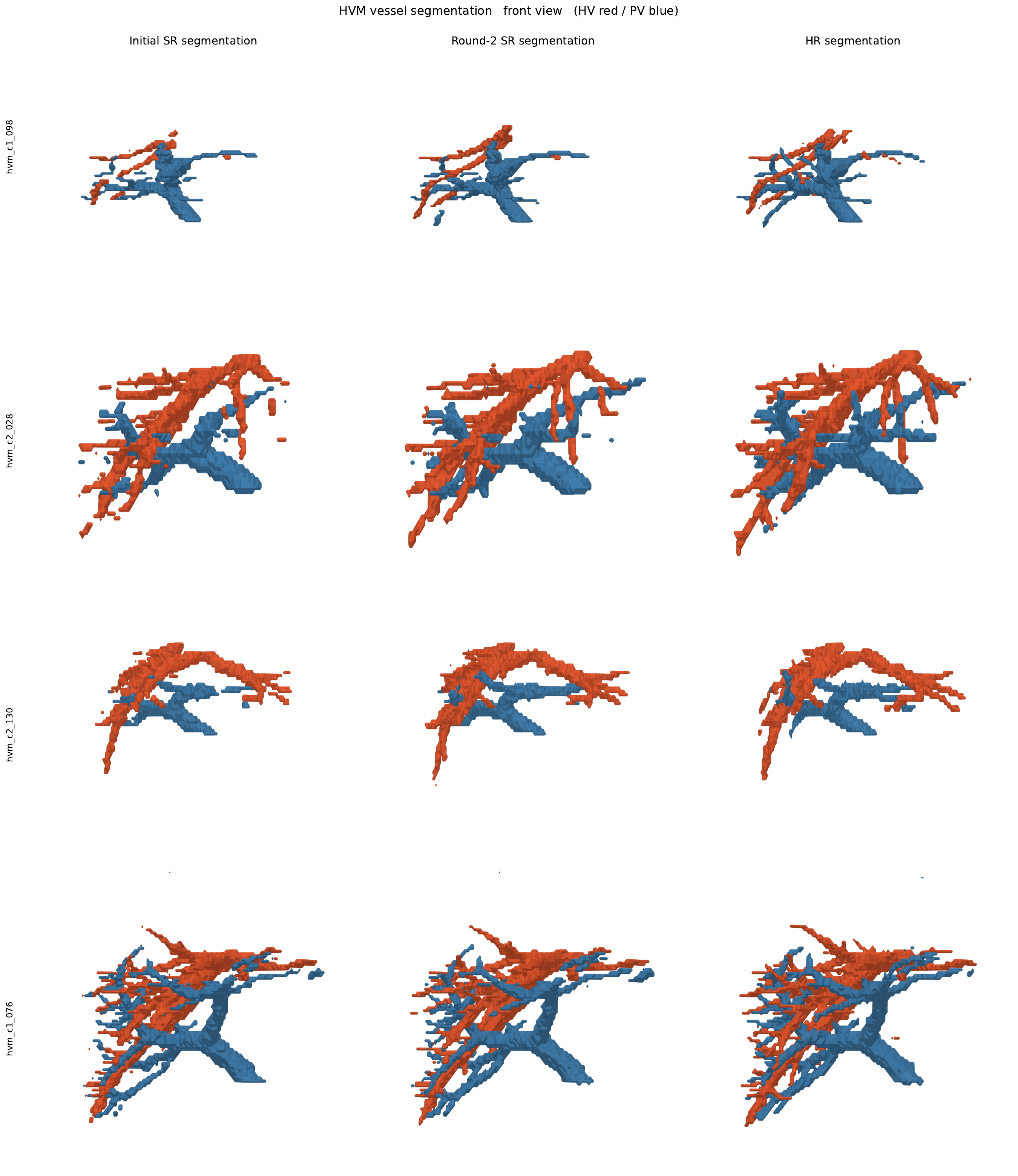}
    \caption{Three-dimensional qualitative comparison on four held-out HVM
    cases. Columns show the initial SR segmentation, round-2 SR segmentation,
    and segmentation of the HR reference. Red and blue denote HV and PV, respectively.}
    \label{fig:hvm-qualitative-3d}
\end{figure}

\clearpage
\section{Why Anatomical Conditioning Rather than Segmentation-Loss Fine-Tuning?}
\label{app:finetuning-comparison}

VascuSR starts from two requirements for vascular CT super-resolution:
reconstruct CT images that are faithful to the
HR reference while preserving vascular anatomical plausibility. A diffusion reconstruction objective addresses the first
requirement, but spatially averaged image agreement alone does not ensure
that the reconstructed vasculature is anatomically plausible. A 3D vessel
representation is therefore a useful source of supervision: it captures
continuity and branching across slices and, with more fine-grained segmentation,
can also provide vessel categories. The design question is how to turn this
structural supervision into a learning signal, and which architecture should
carry it.

The most direct approach is to add Dice and soft-clDice to the SR objective
and fine-tune the image generator through a frozen segmenter. This approach
works: on MSD it improves Dice and clDice over ResShift (Table~\ref{tab:main-results}).
To explain why we use anatomical
conditioning rather than direct fine-tuning, we first compare the two training
strategies, then examine the limitations of this fine-tuning baseline and the
advantages of VascuSR in image fidelity in the vessel neighborhood, vascular structure, and training
cost.

\subsection{Two Ways to Introduce Anatomical Supervision}
\label{app:finetuning-objectives}

We seek an SR image $Y$ with both image fidelity and anatomically plausible
vessels. These requirements can be expressed by the conceptual objective
\begin{equation}
    \min_Y\left[
        \underbrace{\mathcal L_{\mathrm{image}}(Y;X)}_{\text{image fidelity}}
        + \lambda\,
        \underbrace{\Phi(T(Y))}_{\text{anatomical plausibility}}
    \right].
    \label{eq:image-anatomy-objective}
\end{equation}
Here $X$ is the LR input, $T$ extracts the vascular representation from $Y$,
and $\Phi$ penalizes anatomically implausible structure. The two strategies
below differ in how they address these image and anatomy requirements.

\paragraph{Direct fine-tuning with Dice and soft-clDice.}
Fine-tuning implements the two requirements through the joint objective
\begin{equation}
    \begin{aligned}
        \min_\omega\quad &\mathcal L_{\mathrm{image}}(Y_\omega;X) \\
        &+\lambda\left[
            \mathcal L_{\mathrm{Dice}}(T(Y_\omega),V^\star)
            +\alpha\,\mathcal L_{\mathrm{soft\text{-}clDice}}
                (T(Y_\omega),V^\star)\right].
    \end{aligned}
    \label{eq:ft-objective}
\end{equation}
Here $Y_\omega$ is the SR output, $V^\star$ is the task-specific reference
segmentation, and $\omega$ denotes the generator parameters. The image and
anatomy losses update the same generator, with $T$ frozen.

We evaluate checkpoints after 1,200 optimizer updates on MSD
and 2,000 optimizer updates on HVM, using learning rates of $10^{-5}$ and
$5\times10^{-6}$, respectively. In both settings, $\alpha$ is gradually
increased from 0 to $0.5$. Small learning rates are used to preserve the
pretrained reconstruction behavior, while the warm-up introduces skeleton
supervision gradually.

\paragraph{Implicit anatomical regularization through conditioning.}
VascuSR addresses Equation~\ref{eq:image-anatomy-objective} through two
separately supervised tasks: the SR model learns image fidelity from HR,
while the refiner learns vascular correction toward the manual reference
$V^\star$ through its flow loss (Equation~\ref{eq:segflow-loss}). In this
supervised interpretation, write the anatomical penalty as
$\Phi(V)=d(V,V^\star)$, where $d$ is a metric on the shared vascular
representation, such as voxelwise $L_1$. For the refined condition
$\widetilde V=R_\psi(T(Y_0))$ and its conditioned reconstruction $Y_1$.
\begin{equation}
    \Phi(T(Y_1))
    =\underbrace{d(T(Y_1),V^\star)}_{\text{final vascular error}}
    \leq
    \underbrace{d(\widetilde V,V^\star)}_{\text{refinement error}}
    +\underbrace{d(T(Y_1),\widetilde V)}_{\text{transfer error}},
    \label{eq:refinement-transfer-bound}
\end{equation}
Thus, the final vascular error is the anatomical penalty in
Equation~\ref{eq:image-anatomy-objective}. When conditioning keeps the
transfer error small, reducing refinement error tightens the bound on that
penalty. Image reconstruction and vascular refinement therefore address the
two terms through separate supervised objectives, with conditioning linking
them.

\subsection{Image Fidelity near Vessels}
\label{app:ft-regional-fidelity}

We assess whether improved vascular segmentation is accompanied by faithful
CT reconstruction near vessels. Table~\ref{tab:ft-roi-image-quality}
compares final SR images with HR within a 3\,mm dilation of the manual HV/PV
union. SSIM and spatial LPIPS are sampled within the ROI from maps
computed on full, unmasked slices.

\begin{table}[H]
    \caption{Image quality within the 3\,mm vessel ROI on the MSD test set.
    Values are patient-macro means; PSNR is in dB. This comparison uses
    VascuSR R3.}
    \label{tab:ft-roi-image-quality}
    \centering
    \normalsize
    \setlength{\tabcolsep}{7pt}
    \begin{tabular}{lccc}
        \toprule
        Method & ROI PSNR $\uparrow$ & ROI SSIM $\uparrow$
               & ROI spatial-LPIPS $\downarrow$ \\
        \midrule
        ResShift
            & \textbf{34.5933} & \textbf{0.815880} & 0.155418 \\
        ResShift+Dice/clDice
            & 33.0302 & 0.786745 & 0.160075 \\
        VascuSR, round 3
            & 34.4383 & 0.815435 & \textbf{0.147031} \\
        \bottomrule
    \end{tabular}
\end{table}

VascuSR R3 improves all three ROI image-quality metrics over fine-tuning.
For a complementary assessment of absolute CT intensity,
Table~\ref{tab:ft-hu-fidelity} reports MAE, RMSE, and signed bias against
native HR within a 3\,mm dilation of the manual HV/PV union.
For $N$ ROI voxels, let $e_i$ be the SR intensity minus the HR intensity at
voxel $i$. Then $\mathrm{MAE}=N^{-1}\sum_i |e_i|$,
$\mathrm{RMSE}=\sqrt{N^{-1}\sum_i e_i^2}$, and
$\text{signed bias}=\frac{1}{N}\sum_{i=1}^{N}e_i$.

\begin{table}[H]
    \caption{Image fidelity within the 3\,mm vessel ROI on the MSD test set.
    Values are in HU; positive bias indicates overestimated intensity.}
    \label{tab:ft-hu-fidelity}
    \centering
    \normalsize
    \setlength{\tabcolsep}{13pt}
    \begin{tabular}{lccc}
        \toprule
        Method & MAE $\downarrow$ & RMSE $\downarrow$ & Signed bias ($\to 0$) \\
        \midrule
        ResShift & \textbf{21.97} & \textbf{29.61} & \textbf{+1.95} \\
        VascuSR, round 3 & 22.21 & 29.86 & +4.14 \\
        ResShift+Dice/clDice & 26.49 & 34.40 & +12.59 \\
        \bottomrule
    \end{tabular}
\end{table}

Tables~\ref{tab:ft-roi-image-quality} and~\ref{tab:ft-hu-fidelity} show that
VascuSR largely preserves ResShift's regional image quality (PSNR, SSIM,
and spatial LPIPS) and image fidelity (MAE, RMSE, and signed bias), with
modest degradation in some measures. We consider this an acceptable
trade-off for pursuing anatomical plausibility alongside image fidelity,
supported by the structural improvements in Tables~\ref{tab:main-results}
and~\ref{tab:condition-ablation}.
In contrast, ResShift+Dice/clDice worsens all six regional measures,
indicating that direct fine-tuning improves Dice and clDice at the expense
of substantially reduced local vascular image fidelity.

\subsection{Cross-Segmenter Evaluation and Local Appearance}
\label{app:ft-cross-segmenter}
\label{app:ft-qualitative}

On the MSD test set, ResShift+Dice/clDice raises VSNet HV Dice from
ResShift's 0.6526 to 0.7010,
close to VascuSR R3 at 0.7072
(Tables~\ref{tab:main-results} and~\ref{tab:condition-ablation}).
To test whether this gain extends beyond VSNet, we evaluate the same
reconstructions with frozen TotalSegmentator
(TS 2.13.0)~\citep{wasserthal2023totalsegmentator}. Fine-tuning falls below
ResShift, whereas VascuSR R3 improves by 2.71 percentage points
(Table~\ref{tab:ft-ts-dice}). Low absolute scores may partly reflect a
coverage mismatch: the manual GT includes large-caliber extrahepatic venous
segments not captured by these TS predictions.

\begin{table}[H]
    \caption{TS vessel Dice on the MSD test set.}
    \label{tab:ft-ts-dice}
    \centering
    \normalsize
    \setlength{\tabcolsep}{18pt}
    \begin{tabular}{lc}
        \toprule
        Method & TS vessel Dice $\uparrow$ \\
        \midrule
        ResShift & .2912 \\
        VascuSR, round 3 & \textbf{.3183} \\
        ResShift+Dice/clDice & .2803 \\
        \bottomrule
    \end{tabular}
\end{table}

\begin{figure}[H]
    \centering
    \includegraphics[width=\linewidth]{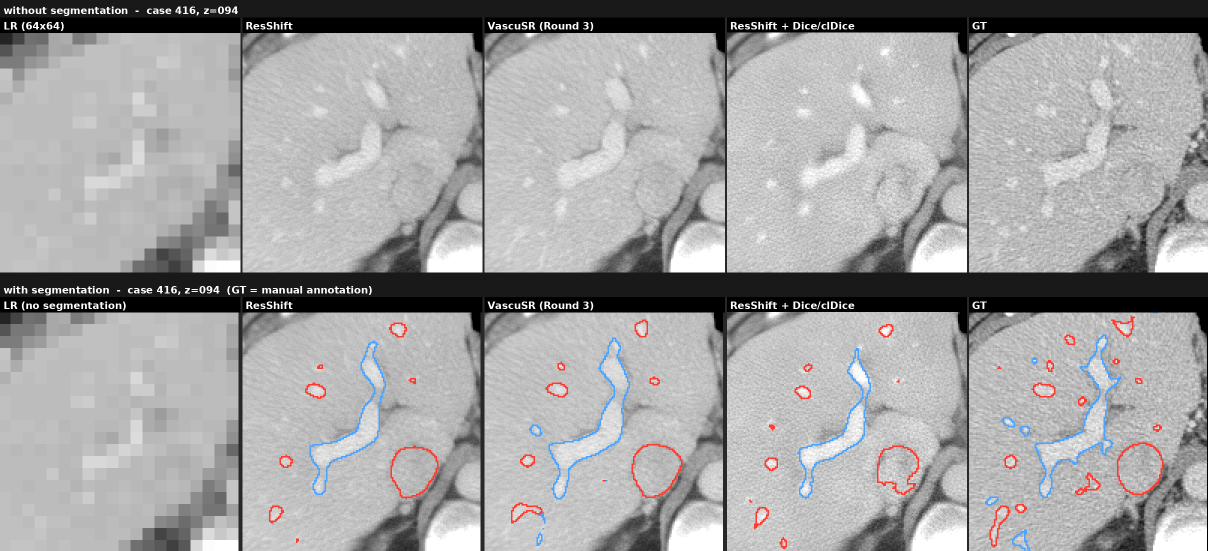}
    \caption{Qualitative comparison on MSD case 416, axial slice $z=094$.
    Columns show LR, ResShift, VascuSR after three feedback rounds,
    ResShift+Dice/clDice, and the HR reference. The upper row shows CT
    images; the lower row adds vascular contours, with HV in red and PV in
    blue. HR contours are manual annotations.}
    \label{fig:finetuning-appearance}
\end{figure}

\begin{samepage}
Figure~\ref{fig:finetuning-appearance} examines the corresponding image
appearance on MSD case 416. VascuSR restores the visible connection between
the two vessel portions in the central PV, whereas fine-tuning does not
visibly improve this connection. In the fine-tuned CT, the vessel portions
are conspicuously bright, while the predicted blue contour still spans their
connection. A segmentation contour can therefore look plausible even where
the underlying intensity profile appears discontinuous.\par
\end{samepage}

The brightening of vessel regions is also supported by quantitative intensity measurements:
ResShift+Dice/clDice has a 3\,mm ROI signed bias of +12.59\,HU, compared with
+1.95\,HU for ResShift and +4.14\,HU for VascuSR R3
(Table~\ref{tab:ft-hu-fidelity}).

When evaluated with TS, fine-tuning shows no Dice improvement and instead
scores slightly below ResShift (Table~\ref{tab:ft-ts-dice}). Together with
vessel brightening and the qualitative intensity--contour mismatch
(Figure~\ref{fig:finetuning-appearance}), this suggests that the model may
have learned a \emph{segmenter-specific appearance shortcut}, making vessels
easier for VSNet to segment without improving their image fidelity.
Its explicit Dice/clDice loss (Equation~\ref{eq:ft-objective}) directly
rewards such segmentation gains. VascuSR instead learns structural correction
in vascular label space and transfers it through anatomical conditioning
(Equation~\ref{eq:refinement-transfer-bound}), while retaining HR-supervised
reconstruction as the image generator's training objective. Without
backpropagating a Dice loss through the generator, this design removes the
direct Dice-based incentive to exploit a segmenter's response and encourages
the generator to use the refined anatomical condition to improve vascular
topology and reconstruction.

\subsection{Vascular Topology}
\label{app:ft-fragmentation}
\label{app:ft-label-space}

Tables~\ref{tab:ft-connectivity} and~\ref{tab:ft-mcd} compare vascular
continuity and centerline accuracy on the same MSD test set. Connectivity
uses 3D 26-neighborhoods.

\begin{table}[H]
    \caption{MSD vascular continuity for the three relevant methods,
    reproduced from Table~\ref{tab:cc-continuity}.}
    \label{tab:ft-connectivity}
    \centering
    \normalsize
    \setlength{\tabcolsep}{7pt}
    \begin{tabular}{lcccc}
        \toprule
        & \multicolumn{2}{c}{HV} & \multicolumn{2}{c}{PV} \\
        \cmidrule(lr){2-3}\cmidrule(lr){4-5}
        Method & \#CC $\downarrow$ & Largest-CC frac $\uparrow$
               & \#CC $\downarrow$ & Largest-CC frac $\uparrow$ \\
        \midrule
        ResShift & 31.0 & .9077 & 26.0 & .8843 \\
        ResShift+Dice/clDice
            & 39.6 & .8936 & 30.4 & .8249 \\
        VascuSR, round 3
            & \textbf{22.7} & \textbf{.9382} & \textbf{22.4} & \textbf{.8931} \\
        \bottomrule
    \end{tabular}
\end{table}

\begin{table}[H]
    \caption{Class-wise MSD MCD values (mm) from Table~\ref{tab:msd-mcd},
    with their HV/PV mean additionally reported.}
    \label{tab:ft-mcd}
    \centering
    \normalsize
    \setlength{\tabcolsep}{12pt}
    \begin{tabular}{lccc}
        \toprule
        Method & HV $\downarrow$ & PV $\downarrow$ & HV/PV mean $\downarrow$ \\
        \midrule
        ResShift & 6.0457 & 8.0877 & 7.0667 \\
        VascuSR, round 3 & \textbf{5.2311} & \textbf{6.6351} & \textbf{5.9331} \\
        ResShift+Dice/clDice
            & 5.6521 & 7.5368 & 6.5944 \\
        \bottomrule
    \end{tabular}
\end{table}

VascuSR improves both connectivity measures over ResShift+Dice/clDice, which even
produces more components than ResShift; MCD also favors VascuSR on both
MSD and HVM (Tables~\ref{tab:ft-mcd} and~\ref{tab:hvm-mcd}).
One possible contributor to the poorer connectivity is the gap between
soft supervision and binary structure. Soft-clDice encourages skeleton
continuity using a differentiable soft-skeleton approximation
\citep{shit2021cldice}. Optimizing this relaxed objective does not directly
enforce every connection in the thresholded mask, so local breaks can remain
after binarization.

Together, the connectivity and MCD results highlight a limitation of direct
fine-tuning: anatomy is specified through manually selected differentiable
penalties (Equation~\ref{eq:ft-objective}). These can favor the chosen scores
while prompting repeated additions of constraints and weight tuning, without
fully capturing what makes a vascular structure anatomically plausible.

VascuSR instead decouples vascular refinement from CT intensity
reconstruction, removing organ texture and other high-frequency image
details from the structural learning task. In this simpler 3D label space,
the lightweight refiner uses a single conditional rectified-flow
velocity-matching loss to learn the mapping from noisy imperfect predictions
to GT (Equation~\ref{eq:segflow-loss}). It thereby learns vascular spatial
dependencies and implicitly models a conditional distribution of plausible
vessel structures from paired annotations, without additional Dice, clDice,
or other morphology-specific losses. The refinement--conditioning loop
transfers this learned anatomical prior to SR while retaining HR
reconstruction as the image objective
(Equation~\ref{eq:refinement-transfer-bound}).

\subsection{Computational Consequences of Decoupling}
\label{app:ft-compute}

Direct fine-tuning is also computationally expensive. Because the segmenter
operates in 3D, each update must generate multiple SR slices, assemble them
into a volume, compute Dice and soft-clDice from the segmentation, and
backpropagate through the segmenter and SR generator. This repeated
computation makes fine-tuning slow in practice.

\begin{table}[H]
    \caption{Training costs on HVM, excluding validation. The refiner is
    trained to the selected 50k checkpoint; ResShift+Dice/clDice is trained
    for 2,000 effective optimizer updates. For VascuSR, we report the
    training cost of the 3D vascular refiner.}
    \label{tab:ft-compute}
    \centering
    \normalsize
    \setlength{\tabcolsep}{12pt}
    \begin{tabular}{lcc}
        \toprule
        Training stage & Wall time (h) & GPU-hours \\
        \midrule
        VascuSR refiner & $\approx9.80$ & $\approx24.60$ \\
        ResShift+Dice/clDice & 29.33 & 94.93 \\
        \bottomrule
    \end{tabular}
\end{table}

Table~\ref{tab:ft-compute} shows that VascuSR's vascular refiner requires
less training time and fewer GPU-hours than direct fine-tuning. Restricting refinement to vascular representation space simplifies
the learning task, allowing a lightweight generative refiner to
learn effective structural correction with substantially less
training time than direct fine-tuning. In addition, training the ResShift backbone jointly with the
anatomy controller takes approximately the same time as training the backbone
alone.

\newpage
\section{Extension to Coronary CTA on ImageCAS}
\label{app:imagecas}

\paragraph{Dataset and evaluation scope.}
We extend our evaluation beyond hepatic vessels to ImageCAS, a coronary
CT angiography dataset~\citep{zeng2023imagecas}. We use a 300-case subset
with 200 training, 50 validation, and 50 test cases. 

\paragraph{Coronary segmenter.}
We initialize from the pretrained Dataset509 coronary model in
TotalSegmentator 2.13~\citep{wasserthal2023totalsegmentator} and fine-tune
the entire network on the 200 training HR CT volumes and their manual
binary coronary annotations. Training uses
$128^3$ patches, batch size 2, and SGD with an initial learning rate of
$0.001$, momentum $0.99$, Nesterov acceleration, weight decay
$3\times10^{-5}$, and polynomial learning-rate decay.

\begin{table}[H]
    \caption{Coronary CTA results on the
    ImageCAS test set. Bold marks the best SR result in each column.}
    \label{tab:imagecas-results}
    \centering
    \normalsize
    \setlength{\tabcolsep}{10pt}
    \begin{tabular}{lcccc}
        \toprule
        Method & Dice $\uparrow$ & clDice $\uparrow$ & PSNR (dB) $\uparrow$ & SSIM $\uparrow$ \\
        \midrule
        RCAN & 0.7699 & 0.8432 & 31.29 & 0.8079 \\
        SwinIR & 0.7702 & 0.8409 & \textbf{31.41} & \textbf{0.8096} \\
        ResShift & 0.7658 & 0.8376 & 29.94 & 0.7647 \\
        VascuSR (Round 3) & \textbf{0.7763} & \textbf{0.8530} & 29.84 & 0.7629 \\
        \midrule
        HR reference & 0.8033 & 0.8610 & $\infty$ & 1.0000 \\
        \bottomrule
    \end{tabular}
\end{table}

\paragraph{Results and interpretation.}
Table~\ref{tab:imagecas-results} shows that VascuSR achieves the highest
Dice and clDice among the evaluated SR methods. The absolute Dice and clDice gains over ResShift in
Table~\ref{tab:imagecas-results} are modest, which may be caused by the fact that coronary vessel
configurations are less complex than those in the
hepatic setting, leaving less room for anatomical correction. Moreover, ResShift already performs close
to the HR segmentation reference. VascuSR closes 28.0\% of the
ResShift-to-HR gap in Dice and 65.8\% of that gap in clDice.
In particular, its clDice reaches $0.8530$, approaching the HR score
of $0.8610$.

\begin{figure}[H]
    \centering
    \includegraphics[width=\linewidth]{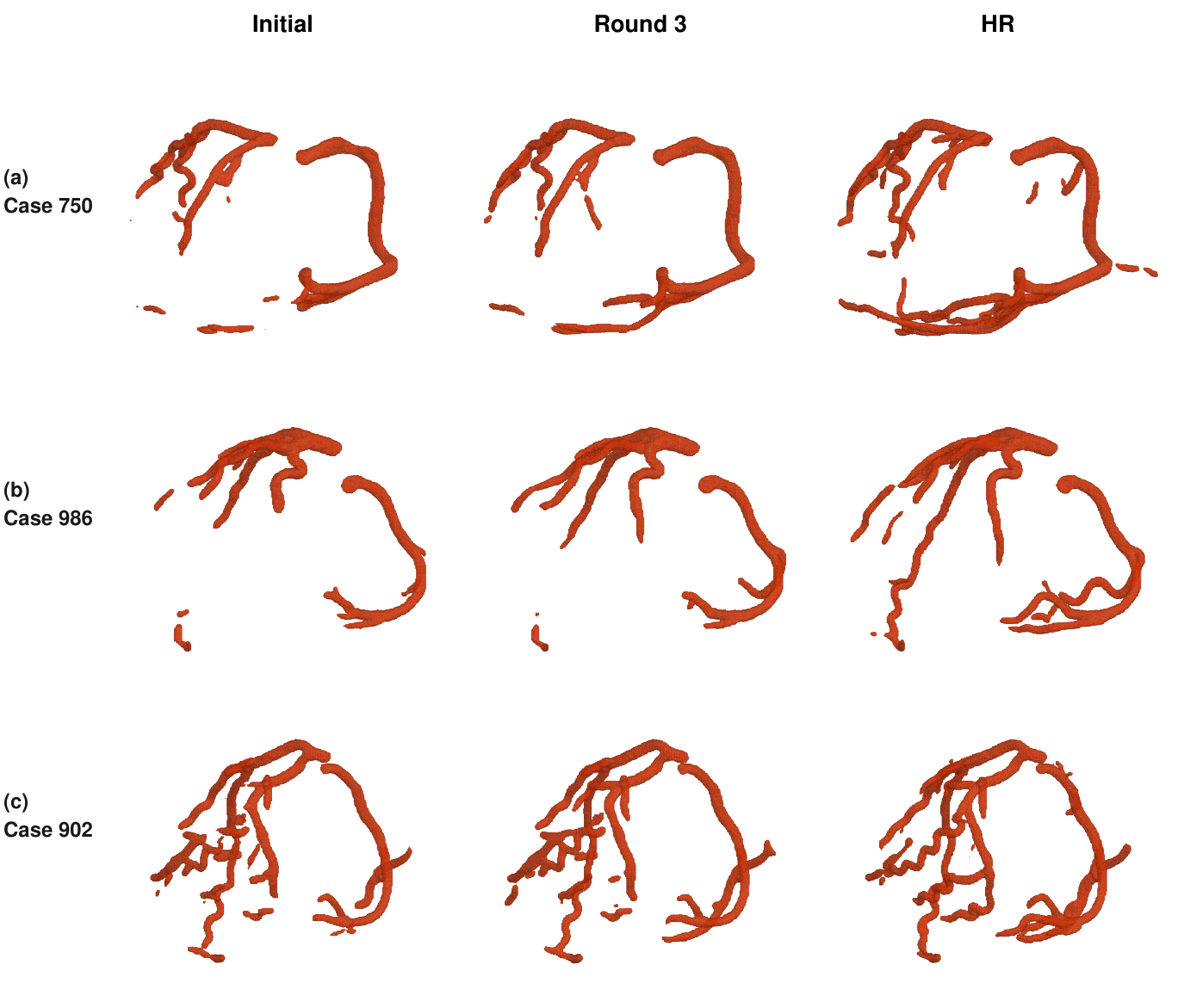}
    \caption{Three-dimensional coronary-artery segmentations for ImageCAS
    cases (a)~750, (b)~986, and (c)~902.
    Columns show Initial (ResShift), Round 3 (VascuSR), and HR
    segmentations.}
    \label{fig:imagecas-qualitative-3d}
\end{figure}

\clearpage
\paragraph{Three-dimensional qualitative comparison.}
Figure~\ref{fig:imagecas-qualitative-3d} compares coronary segmentations
on cases 750, 986, and 902 at $8\times$ in-plane super-resolution.
These examples illustrate improved overlap and centerline agreement. 

\begin{figure}[H]
    \centering
    \includegraphics[width=.94\linewidth]{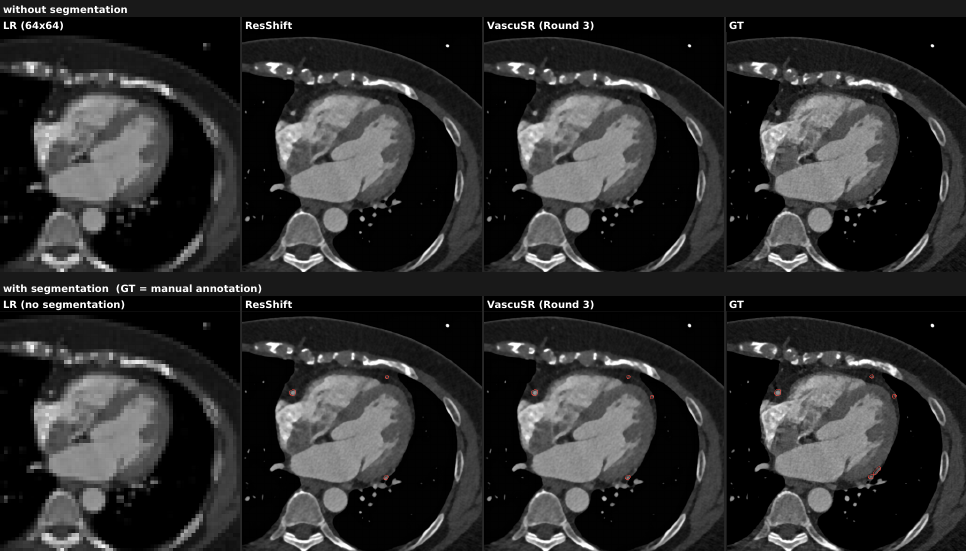}
    \caption{Axial qualitative comparison on ImageCAS case 750 at
    $z=135$ and $8\times$ in-plane super-resolution.
    Columns show LR, ResShift, VascuSR after three feedback rounds, and
    the HR reference. The upper row shows CT images and the lower row
    adds red coronary contours. HR contours are manual annotations.}
    \label{fig:imagecas-qualitative-axial}
\end{figure}

\clearpage
\paragraph{Axial qualitative comparison.}
Figure~\ref{fig:imagecas-qualitative-axial} shows an axial slice from
ImageCAS case 750. In the upper-right region,
VascuSR after three feedback rounds recovers a very small coronary vessel
contour that is present in the manual annotation but missed by ResShift.

\end{document}